\documentclass[conference]{IEEEtran}

\usepackage[T1]{fontenc}
\usepackage[utf8]{inputenc} 
\usepackage{hyperref}
\usepackage{xspace}
\usepackage{enumitem}

\usepackage{amsmath,amssymb,amsfonts}
\usepackage{algorithmic}
\usepackage{graphicx}
\usepackage{textcomp}
\usepackage{xcolor}
\def\BibTeX{{\rm B\kern-.05em{\sc i\kern-.025em b}\kern-.08em
    T\kern-.1667em\lower.7ex\hbox{E}\kern-.125emX}}

\providecommand{\dr}{{DINOS-R}\xspace}
\providecommand{\Sent}{{AUV Sentry}\xspace}
\providecommand{\sent}{{Sentry}\xspace}

\begin{document}

\title{
    Deployment and Development of a Cognitive Teleoreactive Framework for Deep Sea Autonomy
}

\author{\IEEEauthorblockN{Christopher Thierauf}
\IEEEauthorblockA{Deep Submergence Laboratory\\
Woods Hole Oceanographic Institution\\
Woods Hole, Massachusetts, USA \\
christopher.thierauf@whoi.edu}
}

\IEEEoverridecommandlockouts
\IEEEpubid{
\makebox[\columnwidth]{978-1-5386-5541-2/18/\$31.00~\copyright2018 IEEE. In IEEE/MTS OCEANS\hfill}
\makebox[\columnwidth]{}
}

\maketitle

\IEEEpubidadjcol

\begin{abstract}
    A new AUV mission planning and execution software has been tested on \Sent.
    Dubbed \dr, it draws inspiration from cognitive architectures and AUV
    control systems to replace the legacy MC architecture. Unlike these
    existing architectures, however, \dr is built from the ground-up to unify
    symbolic decision making (for understandable, repeatable, provable
    behavior) with machine learning techniques and reactive behaviors, for
    field-readiness across oceanographic platforms. Implemented primarily in
    Python3, \dr is extensible, modular, and reusable, with an emphasis on
    non-expert use as well as growth for future research in oceanography and
    robot algorithms. Mission specification is flexible, and can be specified
    declaratively. Behavior specification is similarly flexible, supporting
    simultaneous use of real-time task planning and hard-coded user specified
    plans. These features were demonstrated in the field on \sent, in addition
    to a variety of simulated cases. These results are discussed, and future
    work is outlined.
\end{abstract}

\section{Introduction}

Scientific uses of AUVs increasingly show the limitations of existing software
for AUV planning, deployment, and runtime. In particular, although the MC
(\textbf{M}ission \textbf{C}ontroller) system in use on \Sent has repeatedly
proven itself for lawnmower patterns, it presents several key limitations
stemming from its rigid implementation. Most notably, it is capable of
executing basic ``go-to'' commands and similar functionality, but was not
engineered for scalability to new mission modalities or real-time
interventions.

As a consequence, MC is unable to perform adaptive sampling surveys, and it
cannot re-task mid-operation based on internal reasoning. For example, if the
on-board CTD indicates a point of interest (such as a hydrothermal plume),
\sent is unable to reason that, because it is ahead of schedule, it should
take an hour to descend and investigate. Human operators also cannot retask
while underway to avoid failure or obtain new objectives. Its behavior is not
interpretable: when erroneous behaviors are taken, operators cannot query
the system as to why. Further, it is not verifiably safe: system bounds cannot
be predicted beyond assumptions stemming from experience. When failures occur,
it cannot
communicate the reasons for these failures and potential solutions, and it
cannot mitigate them.

Expert operators are able to overcome many of these challenges through clever
exploitation of MC's features, or specialized one-off programs which extend it.
MC does permit an acoustic interface to inject a limited set of commands, and
so basic functionality like speed or altitude can be manipulated. However,
real-time retasking of objectives (such as adding new survey zones or adding
new keep-out zones) remains infeasible. Regardless, these workarounds introduce
risk and do not fully resolve these open needs.

The cost is both financial and scientific. Plans take longer to produce and may
be more prone to error, because they require specialized expert operators to
carefully consider each step. Certain deployments are not feasible as a
consequence of poor adaptability and retaskability, and so certain scientific
objectives cannot be obtained. Real-time mission data cannot be acted upon, and
so scientists are forced to budget for multiple exploratory dives rather than
one that explores and then collects high-value data. In the event of failure,
neither the AUV nor its human operators cannot adapt the mission in real time
to correct. There is additional developmental cost: adding new features is
challenging, and supporting new mission modalities is non-trivial.

\begin{figure}[t]
    \centering
\includegraphics[width=\columnwidth]{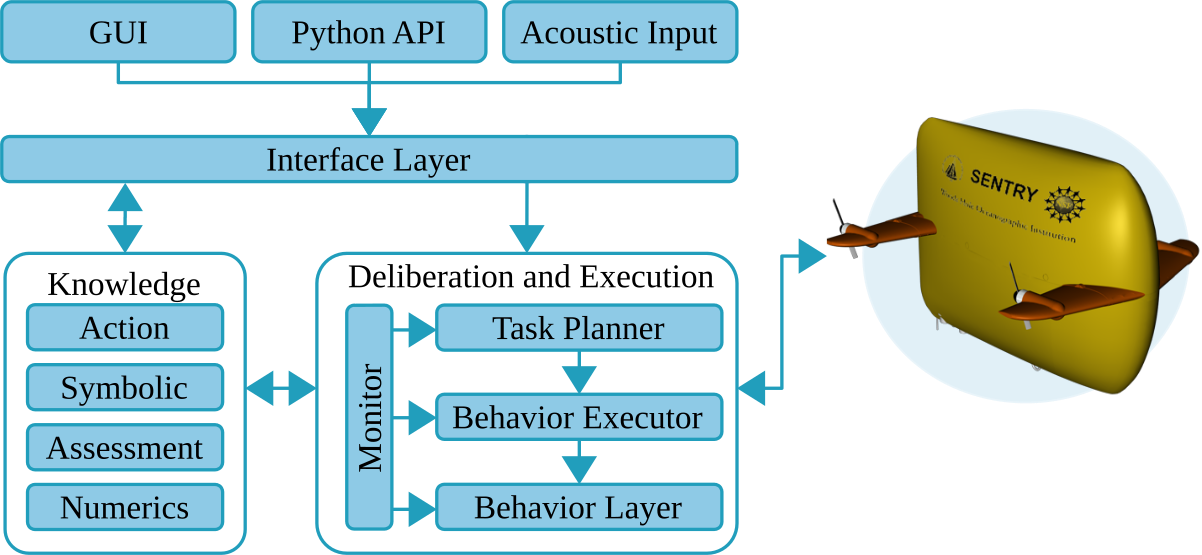}
\caption{\label{fig:architecture}The DINOS-R Architecture. See Section~\ref{sec:architecture} for detail.}
\end{figure}

This motivates the creation of a new mission planning and control architecture
for \sent, dubbed \dr. The key difference between \dr and existing mission
execution architectures is a grounding in cognitive architectures, which have
broadly been successful at producing interpretable and complex long-horizon
robot behavior. Conversely, \dr differs from cognitive architectures by
focusing on AUV mission specification and deployment, which cognitive
architectures have not (generally) been deployed to. In particular, \dr places
a greater emphasis on AUV field-readiness than exploration of novel
architectural components, and does not consider learning and human interaction
as a priority. For these reasons, it should be considered closer to a
teleoreactive architecture than a truly cognitive one, but it draws heavy
inspiration from each.

\dr has several notable features. It is a symbolic orchestrator of reactive
behavior, and so it produces easily interpretable and quickly modifiable
behavior. It contains a long-horizon planning system tightly integrated with a
failure identification system, and so it is capable of mitigating failures
should they occur. Ongoing system decisions and commands occur at a
human-interpretable level, and domain-specific information is provided to the
system using Python to avoid use of niche languages. It provides tooling to add
new behaviors and new implementations of existing behaviors, while ensuring
that these behaviors remain within a framework that performs behavior
reasoning, planning, and assessment.

\dr has been tested in the field, although additional development is required
and remains ongoing. This paper describes these developments. Background in
other architectures is presented, and a set of guiding development principles
is introduced. A field test aboard \Sent is described before discussion of
future work.

\section{Background}

MC is an in-house system that has behaved as a stable workhorse for the Sentry
program across over 700 dives. Expert operators use a standardized set of
Matlab programs to define waypoints, zones of interest, altitude changes, and
other mission steps. These programs then output a series of command strings
which, when provided to MC in sequence, will cause \sent to perform the desired
mission. This allows experts to maintain direct predetermined control of robot
behavior, and performance can be reasonably predicted. When underway, basic
interventions (generally for safety or basic monitoring) can be performed
acoustically.

As previously outlined, however, this does present key limitations. Adaptive
sampling with Sentry has only been performed using highly specialized programs
which do not generalize to other domains. Missions must be specified by hand by
expert operators, locking the user out of contributing directly to mission
planning and increasing the potential for operator error. Further, it
introduces greater development cost when new behaviors or hardware are to be
integrated. Perhaps most critically, MC is not resilient: though it has proven
reliable when within scope, the fact that it operates by naively following
predetermined steps prevents it from responding to changes in the environment
or capabilities. Sentry faces these limitations despite considerable advances
in robot control architectures.

Reactive architectures are appealing for their ability to address the
resilience problem. Most famously, subsumption~\cite{Brooks1986Subsumption}
demonstrated robust low-level behaviors, but lacked explicit mechanisms for
deliberative planning. This limitation prevents working towards real-time
goals, a challenge that has been addressed using three-layer architectures (as
a non-comprehensive listing, consider SFX~\cite{Gat1998SFX},
AuRA~\cite{Arkin1990AuRA}, or ATLANTIS~\cite{Bonasso1997ATLANTIS}). By
dedicating layers to deliberation, sequencing, and reactive execution, fairly
robust behaviors can be maintained alongside long-horizon sequencing of said
behaviors.

Cognitive architectures like DIARC \cite{Scheutz2007DIARC}, Soar
\cite{Laird2012SOAR}, ACT-R \cite{Anderson1996ACTR}, RCS \cite{Albus1991RCS},
and others have advanced beyond these systems through an emphasis on
reasoning, communication, and learning by seeking to be as human-like as possible.
DIARC, in particular, has demonstrated that it can integrate with a variety
of platforms, reason about failure and future actions, and conduct dialogues.
These dialogues permit human instruction of complex tasks alongside robot
explanations of ongoing failure or failure mitigation strategies.
While we do not require the ability to speak with
\sent while it operates sub-sea, this capability highlights the potential
to interpret and instruct behaviors of these systems, which is highly desirable.

Despite the impact of cognitive architectures on domains like human-robot
interaction and unmanned field vehicles, these realizations have broadly not
made their way to oceanography. That said, deliberative control
systems have played a crucial role in enabling long-horizon missions.
For example, T-REX~\cite{McGann2008TREX}, the ``\textbf{t}eleo\textbf{r}eactive \textbf{ex}ecutive'', provided the ability to
sequence reactive behaviors using a temporal task planner, and was
designed specifically for AUVs. However, it required specialized knowledge
(in particular, of the NDDL language) that has prevented use outside
of niche operators, and as such it is no longer maintained. More
widespread architectures, like MOOS-IvP~\cite{Benjamin2010IvP} are
capable of mission-level autonomy, but are generally very rigid
and do not perform long-horizon planning, failure analysis, or re-tasking.

Aerospace mission planners (e.g.,~\cite{Frank2001Europa, Rabideau2019MEXEC})
provide valuable insights in automating temporal planning for long-horizon
safety-critical missions, but are often so safety-critical that mission
planning requires a high element of human--in--the--loop. Additionally, they
often depend on an operational loop of human planning, robot acting, then
waiting in a safe state while another set of instructions is crafted. This
approach is not viable sub-sea, where sea currents and other dynamic factors
require agents to constantly adapt to their environment.

There are works providing general-purpose real-time planning systems. Notably,
ROSPlan~\cite{Cashmore2015ROSPlan} provides a planning framework, already
integrated into the ROS middleware~\cite{quigley2009ros} (which is currently
in use on Sentry), and remains both robot and domain agnostic. Other
architectures (consider CLARAty~\cite{Nesnas2006CLARAty} or
FogROS~\cite{chen2021fogros}) offer similar features. However, it does not
provide the tight coupling between deliberative and reactive behaviors that we
require here, does not provide reasoning about failure and similar resilience
strategies, and does not provide tools for non-experts to interpret and produce
complex robot behavior.

To prioritize interpretability by non-expert operators, alternative strategies
are often presented. Perhaps most notably, behavior trees~\cite{Colledanchise2018BTs}
provide visual and often intuitive representations of behavior, while potentially remaining
modular and reactive. They remain a sensible choice for many applications, but are
not suitable here. They struggle to scale, particularly for complex missions; they
cannot be re-tasked or provided with new information, states, or behaviors; and they
lack temporal reasoning and other features. Like finite state machines, they require
substantial scaffolding to support the functionality already present in cognitive
architectures, though they remain useful as a strategy that these architectures can employ.

Thus, merging the ease-of-use of an explicitly interpretable cognitive
architecture with the functionality and reliability of reactive system provides
a compelling path for complex mission deployments. This work builds upon the
power of Cognitive Architectures (most notably DIARC~\cite{Scheutz2007DIARC})
as well as the success and shortcomings of three-layer and teleoreactive
frameworks (most notably, T-REX~\cite{McGann2008TREX}). In doing so, an
architecture is produced which is interpretable and extensible for AUV mission
planning, and is capable of robust, long-horizon planning, real-time retasking,
and adaptive and resilient behavior.

To accomplish this, \dr provides a toolkit for planning and executing missions
by providing components for real-time long-horizon planning, trackline
generation, runtime monitoring, and self-assessment. Additional hooks for
planned components are also provided. This produces an architecture that is
capable of running standard ``lawnmower'' missions as well as complex real-time
retasking and adaptive sampling, while ensuring that it is capable of growth
and adaptation to other platforms. The specific components and rationale that
enable this are outlined next.

\section{Design Philosophy}

\sent is used in a dynamic range of deployments: although all deployments can
be summarized as ``go to points of interest and collect data there'', the
strategy used to obtain each point (e.g., lawnmower vs.~series of go-to's or
some combination of both) and the data collected at each (e.g., multibeam SONAR
vs.~photography vs.~sidescan SONAR) varies deployment to deployment. Thus,
the software which \sent employs must be similarly dynamic.

At odds with the need for complex dynamism is the need for simplicity and
interpretabiliy. \sent is deployed by technical experts at the behest of
clients who are (generally) scientific experts with unique scientific needs.
Neither of these are familiar with niche languages like LISP or NDDL, and this
lack of familiarity cannot be an obstacle to their use of the system.
Therefore, behavior must be orchestrated at a high-level which is easily
human-interpretable.

\sent has a unique opportunity to provide software which remains broadly useful
to researchers beyond just the \sent program. Additionally, \sent is frequently
swapping out new components and functions, and so any mission controller used
by \sent must be fairly modular: portions of the architecture should be
easily re-configured and swapped in or out for experimentation or integration
with new devices. This also drives the interest in developing such a system
in-house and pushing it as an open-source release.

Despite this modular approach, \dr will strive to be monolithic rather than
component-driven. Though other architectures consider this to be
disadvantageous, for \dr it is an explicit goal. A monolithic architecture
intentionally prioritizes simplicity, ease of use and understanding, and ease
of integration and debug. Though distributed architectures offer more
theoretical advantages, for field readiness these advantages are
outweighed by simplicity and risk minimization, and a monolithic architecture
remains capable of producing the behavior required of AUVs.

\sent is an operational platform, not an experimental one. This drives the interest
in a monolithic architecture, but more broadly presents the primary motivator.
Any architecture produced for \sent must ultimately be highly reliable and be
reasonable for a team of engineers to maintain, while producing useful behavior
for scientific data collection sub-sea.

\section{Architecture}
\label{sec:architecture}

\dr is a toolkit for producing simple or complex robot behavior. It provides
library calls prior to deployment to describe, assemble, specify, and analyze
robot behavior or system outcomes. Core functionality is implemented as plugins
that are provided to a central orchestrator. This allows any new implementation
of each core function to be provided for experimentation or custom
configurations.

During runtime, an execution cycle that performs and self-monitors ongoing
behaviors may re-task as appropriate (depending on safety or operator-specified
triggers). Throughout the mission, \dr continuously selects the most
appropriate behavior and evaluates its status in real time. This selection and
evaluation is based on mission-specified goals or instructions, as well as
system-specified safety parameters. These are defined by extending Python as a
domain-specific language in the hopes that this reduces the bar of entry for
non-experts.

Note here also another difference between \dr and other mission planning
systems: \dr does not draw a distinction between the planning and execution
phase. Anything that can be performed on land can be instructed to the system,
in real time, while underway subsea. This dramatically improves adaptability
and functionality.

Several threads run in parallel, producing a monolithic architecture that
remains lightweight and middleware-agnostic. When a behavior is run, it is run
independently; the self-monitoring and self-assessment processes also run
independently. This ensures that no middleware is a dependency, but behaviors
which are being called may still depend on them. As a result, either ROS 1 and
ROS 2 can be used, or neither if this is preferred. Though this monolithic and
strictly structured approach presents some limitations, they are balanced out
by the behavior construction system and provide key advantages stemming from
simplicity: most notably, ease of engineering and debug in--the--field.

The core of \dr is written in Python~3.
There are several core systems which can be swapped out as necessary, assuming
they meet system-specific interfaces.
Behavior specifications are outlined in Python: PDDL-like information about
what conditions must be met before and after a behavior can be provided
programmatically. Larger sets of behaviors can be assembled using Python's
import system, allowing complex vehicle-specific or mission-specific sets of
behaviors to be quickly assembled.

The implementation and definition of behaviors are strictly separated. This is
a concept borrowed from DIARC~\cite{Scheutz2007DIARC} that has proven valuable both
developmentally (the system does not require knowledge of the hardware to be
run on it, and any system which implements the appropriate interfaces can
execute a plan) and conceptually (the system can reason about behaviors it does
not yet have). Python function annotators are provided, allowing the programmer
to indicate what functions should be associated with what behaviors.

Behaviors can also be constructed using other behaviors. This permits the
integration of behavior trees, reinforcement learning, finite state machines,
or other techniques. Because these produce behaviors, and behaviors can be
constructed into other behaviors, it is reasonable to expect that developers
will produce complex core behaviors using these tools, that mission planners
will produce missions using simple sequences, and operators will inject new
steps to the plan during execution as needed.

For deployment, \dr is provided with a series of core systems, each implemented
as a plugin. A \dr agent is four central databases that track the system and
its capabilities, the deliberator, and robot-specific implementation classes.
Because the individual functions which implementation robot behavior are tagged
with relevant behavioral information, the deliberator learns its facts about
how behaviors interact with the environment as it consumes each class it can
employ. However, these facts can also be provided absent of any implementation.

The databases centralize information to prevent duplication and to provide
common access, which simplifies debug and the introduction of new systems.
Again borrowing from DIARC~\cite{Scheutz2007DIARC}, there are several which
enable high-level reasoning. First is the behavior database, which stores
information about what behaviors are available, their preconditions and
effects, and how they can be called. The second is a belief database, which
tracks symbolic states which are believed to be true about the environment.
Some of these facts are inferred by the success of completing a behavior (e.g.,
we infer that going to a location means we are now in that location) while some
are explicitly observed (e.g., by referencing deadreckoning data). Noting a
mismatch in these becomes a valuable self-checking behavior. Third is a
`numerics' database, which associates high-level symbolic information (``survey
zone a'') with concrete data (a set of coordinates). This separation simplifies
the planning stages and allows behavior reuse: planning occurs in the abstract
(``I will need to survey zone a'') and then behavior implementations are
acquired as necessary (``to survey zone a, I first will go to the
coordinate...''). Fourth is the self-assessment database, which tracks
behaviors over time (using metrics like time, battery use, success percentage)
for real-time self-checking and future planning.

The deliberator monitors current goals. These goals are high-level
symbolic goals, not control system goals (e.g., ``survey this area'' rather
than setting a velocity target). By default, \dr prioritizes user-specified
behavior but will switch to self-directed behavior when needed. Plans and
behaviors are constructed using symbolic representations (e.g., “the survey
zone”, “the start location”) rather than explicit numerical values. The system
automatically associates these symbols with concrete values at runtime. This
reduces cognitive load on mission designers and minimizes human error.
This improves reusability and allows symbolic references to be reassigned or
overridden at runtime.

As behaviors are run, they are monitored by the deliberator. Global
safety parameters (such as maximum depth or keep-out zones) are actively
monitored, and behaviors will be halted in favor of self-preservation
behaviors if necessary. Behaviors may also be halted when some behavior-specific
termination condition is met (e.g., a ``go to waypoint'' behavior is halted when
we are observed to be within a reasonable range of that target).

\section{Deployment}

These tests took place through the NSF support for engineering work on
\Sent and HOV Alvin. Testing occurred aboard the R/V~\textit{Atlantis} in early March~2025
on \sent dive 768.

The first set of tests took place as a digital twin setup. Throughout the
at-sea time for this testing period, \sent was deployed for other objectives
using the standard MC approach. In parallel, \dr was made to replicate these mission
deployments in a physics simulation\footnote{Simulation testing was performed
using an environment created in the Gazebo~\cite{koenig2004design} suite.}, and
real-world mission behavior was compared to simulation behavior. This increased
confidence prior to actual deployment.

The objective of the real-world deployment was primarily to validate simulation
performance. Secondary objectives were to build confidence in a real-time task
planning system while at-depth, and to show the system is capable of performing
pre-scripted missions. To that end, two approaches for behavior selection were
used. First, a set of high-level goals were provided with no instructions on
how to solve them, showing the system can work towards complex goals
autonomously.

A high level goal was provided. This plan involved full coverage of a
predefined region. The system was informed that by the end of the mission, it
should have completed a survey at the appropriate depth for a multibeam survey
and dropped all weights\footnote{Note that \sent enters a deployment
over-ballasted, dropping a first set of weights to become approximately
neutrally buoyant for deep-sea operation, before then dropping the remainder of
weights for ascent.}. This goal was specified using a custom domain-specific
language implemented in Python3 in two parts. The first part was a symbolic
description of the goal (e.g., {\tt did\_survey(zone\_a)} $\land ${\tt
at\_depth(surface)}). The second portion was information about these symbolic
operators that was essential for execution (e.g., the specific coordinates that
define {\tt zone\_a}).

When provided to the task planer, this automatically generated a plan with the
appropriate steps. The system appropriately determined that descent,
self-calibration, survey tracklines, and ascent would be necessary in that
order. Then, the actual paths for tracklines and connections between relevant
waypoints were generated. The descent, calibration, and ascent behaviors are
pre-existing and fairly hardcoded, but the survey tracklines are not: they must
be computed for full coverage of a specific zone.
This full-coverage computation was performed automatically, and the new
behavior satisfying the coverage requirement was added to the behavior system.
During initial deployment, the generated pathing was replaced with pathing from
a previous dive for predictability and safety. This replacement was done while
the vehicle was idling in a fully-running, ready-to-deploy state.

For purposes of predictability and safety during this initial deployment, the
behavior which executes this generated pathing was replaced with a behavior
which executes a pathing from a previous dive. This also occurred while the
vehicle was idling in its fully-running, ready--to--deploy state. In doing so,
we observe that actions can be generated, modified, and made available
dynamically, even while in operation.

The descent behavior, like the pathing behaviors, is actually more complex than
a simple hard-coded event. A series of events must occur in a specified order:
first, the actuated fins must be set to the appropriate orientation for
descent; then the system must wait while providing status information until
within an appropriate range to finish the descent sequence; then it must wait
until the DVL\footnote{The Doppler Velocity Log, while generally used for
    measuring speed and direction over ground, also provides an altitude
measurement that we use to ensure we are within reasonable range of the sea
floor before dropping weights and finishing the descent sequence.} shows
appropriate altitude; then vertical thrust is initiated to slow descent before
the descent weights are dropped and operation begins. Rather than hard-coding
this functionality, this behavior is composed of several other behaviors (set
thruster position, wait until depth, etc). This eases behavior implementation,
increases reliability through reuse, and eases modification for future
missions.

This has already demonstrated both on-deck generation of behavior sequencing as
well as the ability to generate and add behaviors dynamically, as appropriate.
However, it is often necessary for operators to introduce strictly-defined
steps in a plan. To test operator overrides of behavior, this goal was provided
alongside a simpler human-produced sequence. The task planning system had correctly
inserted a magnetometer calibration step prior to the survey based on system
knowledge that magnetometer use depends upon calibration. For this deployment,
however, this is non-critical, and a new plan was inserted which removes that
step. This demonstrates on-surface operator overrides. It also shows that although
the task planner remains central to anticipated operation, it is possible to deploy
a more typical hard-coded mission plan, either in parallel or independently.

The dive began as a fairly standard \sent deployment, with typical deployment
procedures followed. Other hardware was being tested on-board in parallel to
this test, but this testing did not impact the deployment. As a result, \sent
was configured with the typical drop-weight configuration for descent, and with
the typical sensor suite (which includes multibeam, sidescan, CTD, and other
sensing).

During deployment, this specified plan was appropriately followed. With the
exception of some implementation issues that are not relevant to the
architecture (most notably, re-implementation of some central behaviors led to
coordinate frame issues), the plan and survey were completed successfully.

The ability to inject new high level goals was also tested. 30 minutes prior to
the anticipated end of the survey, \sent was instructed to abort the mission.
Importantly, this was not done using a specialized abort behavior (although one
does exist). Instead, an acoustic command was sent that modified the current
goal. This new goal was that \sent must be ready for recovery. As a response,
\dr produced a new sequence of behaviors (dropping weights, changing thruster
orientation, waiting for recovery), and prioritized these because they
originate from the operator. This showed that new goals can be inserted and
prioritized, while ensuring safe recovery and concluding the dive.

Recovery completed without incident and was otherwise standard.

\section{Discussion and Future Work}

The results of these in-simulation tests and real-world deployment show promising
results with much future work. Most notably, it has demonstrated that a symbolic
planning system is capable of providing equivalent functionality to hand-crafted
plans aboard \Sent, while also showing that it can exceed current capabilities.
Additional development is ongoing in several key areas, with future developments
still being planned.

First, access to non-expert users is being prioritized. A web-based GUI is in
development which will permit science users to perform their own mission
planning, and evaluate outcomes and objectives, while using the same tooling
that the expert operators will employ. This changes the relationship between
\sent, the \sent operator, and the science user: rather than a
client--operator--service relationship, the client and operator will
collaborate with \sent to produce a mission that meets science goals alongside
safety needs. Similarly, because this approach allows programmatic control of
complex behavior, an API is in development to allow the science user to have a
``backseat driver'' program. This will allow the scientist to direct behavior
of \sent using their custom algorithms, while the \sent team remains confident
that behavior is constrained to safe operation.

Related to these is an open-source release and extension to other AUVs. The
core problems being addressed here are broadly applicable to other autonomous
undersea systems, and much duplicate effort is spent on creating similar
systems of autonomy. Ideally, this architecture (or a similar one) can form
the foundation of a broader collaboration among users of scientific AUVs,
where generally-useful behaviors or functions can be shared.

Other features are also in development. Preliminary work has integrated linear
temporal logics (LTLs) into the planning system and GUI, so that basic safety
guarantees (e.g., ``never enter this zone'', ``always finish with greater than
$x$ battery percentage'') can be enforced. Integrations with the
PRISM~\cite{hinton2006prism} probabilistic verification system and
SPIN~\cite{holzmann2011model} model checker are currently being tested. Multi-agent
deployments have not yet been performed using \sent, but this approach supports
such a development and so this is being considered. Finally, additional tooling
to support ease of real-time use and introspection is being developed.

\section{Conclusion}

\dr, a novel architecture for AUV mission planning and deployment, was
developed and first tested on \Sent. Although it remains in development,
it seeks to replace the legacy MC system by allowing safely retaskable
behavior. By adopting techniques from the cognitive architecture space
that have not yet been brought to deep-sea autonomous systems, \dr has the
potential to enable more robust deployments and provide access to scientific
data collection that were previously inaccessable. These claims were validated
in a preliminary field deployment on board \sent.

\bibliographystyle{IEEEtran}
\bibliography{robot-architectures}

\end{document}